\documentclass[letterpaper, 10 pt, conference]{ieeeconf}  

\IEEEoverridecommandlockouts                              

\usepackage{cite}

\usepackage{amsmath,amssymb,amsfonts,amsthm}
\usepackage{algorithmic}
\usepackage{algorithm}
\usepackage{graphicx}
\usepackage{textcomp}
\usepackage{xcolor}
\usepackage{svg}
\usepackage{dsfont}
\usepackage{adjustbox}
\usepackage{xspace}

\newcommand{\transpose}[1]{#1^\mathrm{T}}

\newcommand{\dd}{\mathop{}\!\mathrm{d}}

\providecommand{\FullStop}{\text{~\@.\xspace}}
\providecommand{\Comma}{\text{~,\xspace}}

\DeclareRobustCommand{\vec}[1]{ 				
	\ifthenelse{\equal{#1}{\omega} \OR \equal{#1}{\varphi} \OR \equal{#1}{\alpha} \OR \equal{#1}{\beta} \OR \equal{#1}{\chi} \OR \equal{#1}{\delta} \OR \equal{#1}{\varepsilon} \OR \equal{#1}{\phi} \OR \equal{#1}{\epsilon} \OR \equal{#1}{\gamma} \OR \equal{#1}{\eta} \OR \equal{#1}{\iota} \OR \equal{#1}{\kappa} \OR \equal{#1}{\lambda} \OR \equal{#1}{\mu} \OR \equal{#1}{\nu} \OR \equal{#1}{\pi} \OR \equal{#1}{\theta} \OR \equal{#1}{\vartheta} \OR \equal{#1}{\rho} \OR \equal{#1}{\sigma} \OR \equal{#1}{\varsigma} \OR \equal{#1}{\tau} \OR \equal{#1}{\upsilon} \OR \equal{#1}{\xi} \OR \equal{#1}{\psi} \OR \equal{#1}{\zeta}}{
		\boldsymbol{#1}
	}{
		\mathbf{#1}
	}
}

\AtEndOfClass{}

\title{\LARGE \bf
Task-Space Constrained Stochastic Trajectory Optimization for Time-Optimal Forestry Crane Motion Planning
}

\author{Marc-Philip Ecker$^{1,2}$, Christoph Fröhlich$^{2}$, Bernhard Bischof$^{2}$, Wolfgang Kemmetmüller$^{1}$, Tobias Glück$^{2}$
\thanks{$^{1}$Marc-Philip Ecker and Wolfgang Kemmetmüller are with the
Automation \& Control Institute (ACIN), TU Wien, 1040 Vienna, Austria
        {\tt\small \{ecker,kemmetmueller\}@acin.tuwien.ac.at}}%
\thanks{$^{2}$Marc-Philip Ecker, Christoph Fröhlich, Bernhard Bischof  and Tobias Glück are with the Center for Vision, Automation \& Control,
AIT Austrian Institute of Technology GmbH, 1210 Vienna, Austria
        {\tt\small \{marc-philip.ecker,christoph.froehlich, bernhard.bischof,tobias.glueck\}@ait.ac.at}}%
}

\begin{document}
\maketitle
\thispagestyle{empty}
\pagestyle{empty}

\begin{abstract}
Efficient, collision-free, and time-optimal motion planning is a fundamental requirement for autonomous forestry cranes operating under hydraulic pump-flow constraints. The Via-Point-based Stochastic Trajectory Optimization (VP-STO) algorithm has demonstrated near-time-optimal hybrid motion planning in this domain, but requires a fixed terminal joint configuration specified prior to optimization. For kinematically redundant manipulators such as forestry cranes, this pre-commitment to a single inverse kinematics solution restricts the planner's ability to exploit redundancy, particularly under the nonlinear, globally coupled pump-flow constraint where admissible joint velocities depend on their combined hydraulic demand. This paper presents TSC-VP-STO, a task-space-constrained extension of VP-STO that replaces the strict terminal joint-space constraint with a task-space constraint, jointly optimizing the trajectory and the redundant degrees of freedom of the terminal configuration. This enables the planner to adapt end configurations to the environment-dependent motion and hydraulic flow allocation, yielding more balanced pump utilization and shorter trajectory durations.
We formalize the approach through a configuration space decomposition and derive a concrete reachability constraint for the forestry crane kinematics. Experimental evaluations across multiple planning targets and via-point configurations demonstrates a reduction on trajectory durations by 12-15\% on average and improved pump-flow utilization compared to the baseline VP-STO. The practical applicability of TSC-VP-STO is validated through real-world deployment on a forestry crane, including a full log-loading cycle.
\end{abstract}
\section{Introduction}
Forestry cranes are among the most demanding platforms for autonomous manipulation research. Operating as large-scale hydraulic manipulators in cluttered, unstructured outdoor environments, they must execute fast, collision-free motions while respecting tight actuation constraints imposed by a shared hydraulic pump. At the same time, the forestry sector faces a growing shortage of skilled operators due to demographic shifts and the high cost of training~\cite{morales:2014,ayoub:2024}, driving strong industrial demand for autonomous and semi-autonomous crane systems. Meeting this demand requires motion planners that are not only safe and collision-free, but that also fully exploit the machine's actuation capabilities to minimize cycle times.

\begin{figure}[t]
\centering
<\includegraphics[scale=0.28]{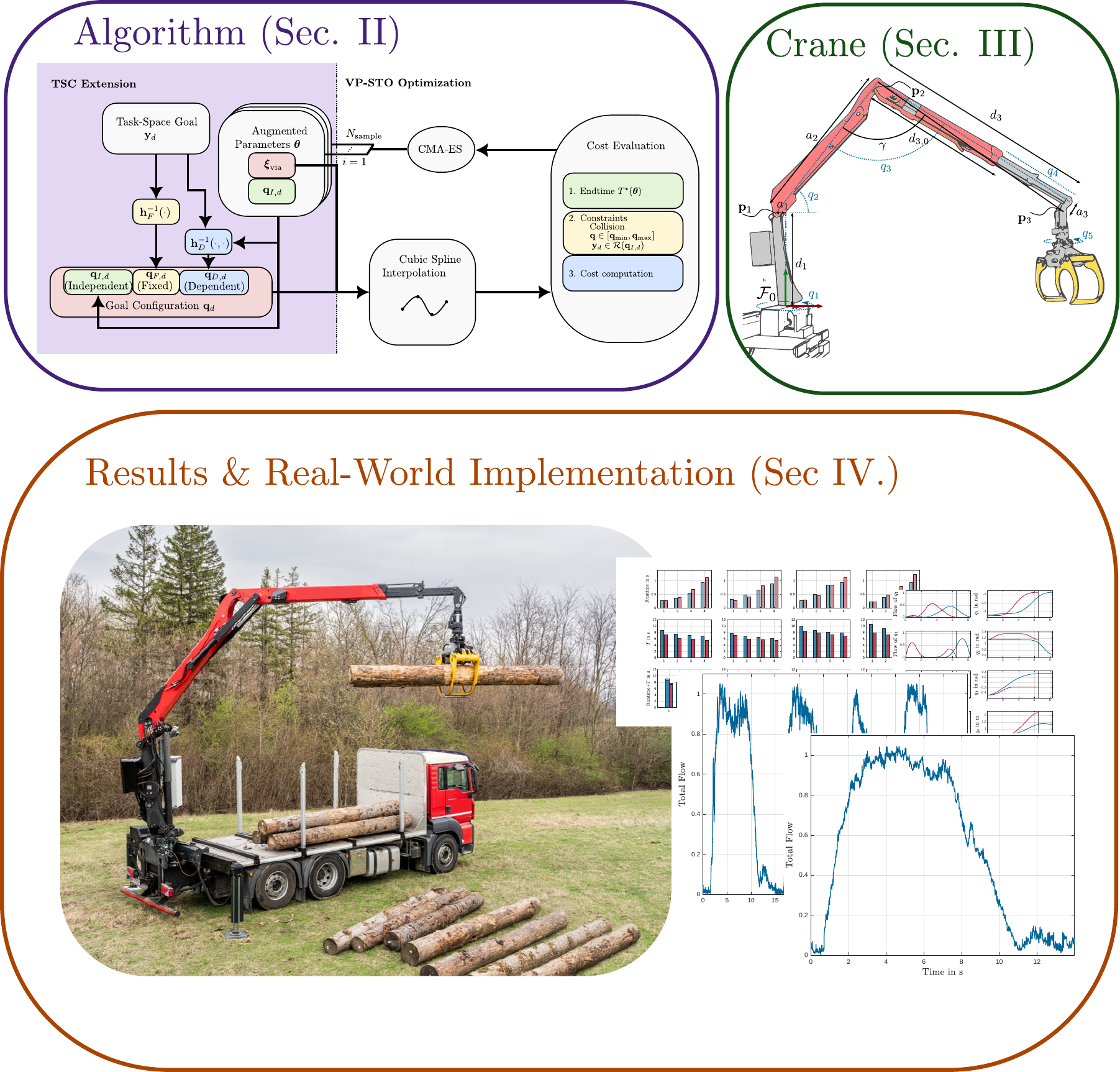}
\caption{Overview of TSC-VP-STO. The proposed method replaces the fixed terminal joint-space constraint of VP-STO with a task-space constraint, enabling the optimizer to jointly refine the trajectory and the redundant terminal configuration for improved hydraulic flow allocation.}
\label{fig:TSCVPSTOConcept}
\end{figure}

\begin{figure*}
\centering
\includegraphics[scale=0.28]{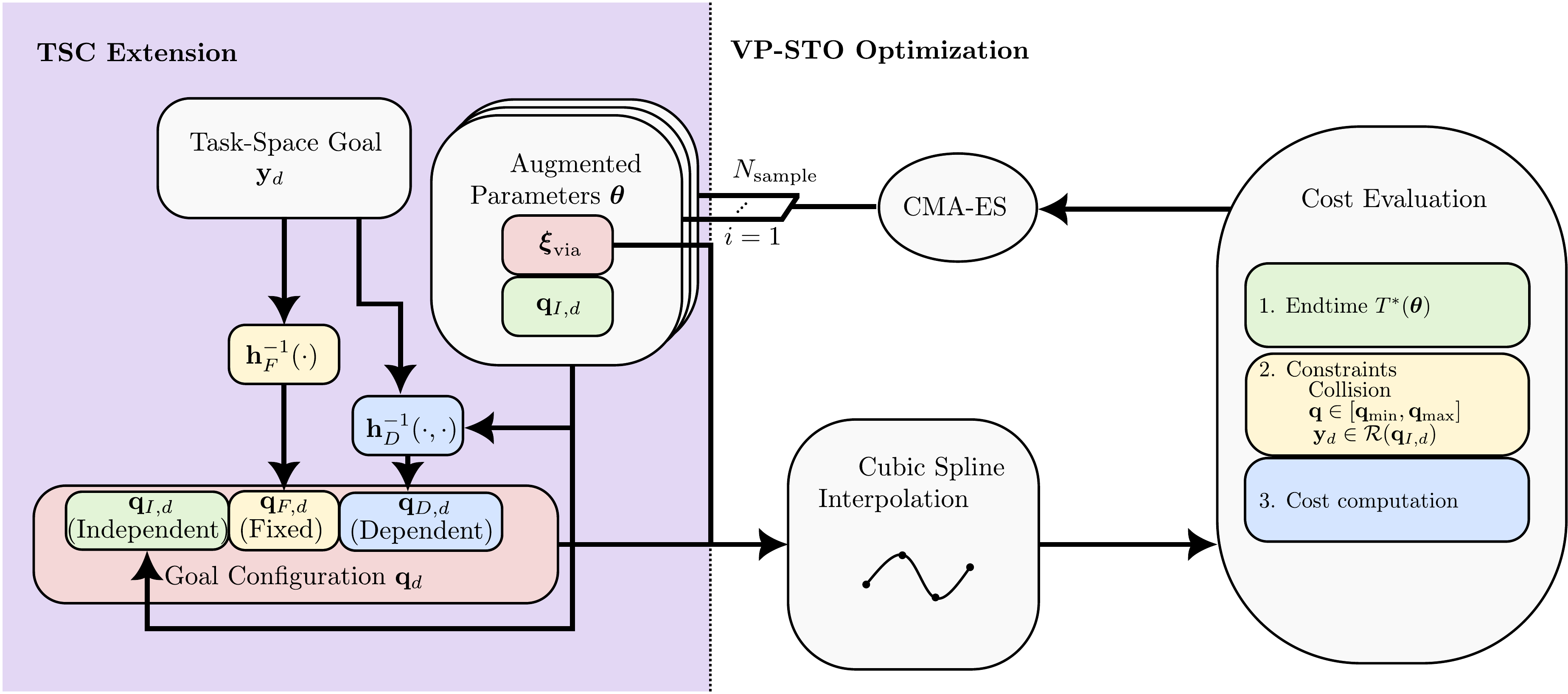}
\caption{Block diagram of the proposed TSC-VP-STO algorithm. The augmented parameter vector $\vec{\theta}$ includes both the via-points $\boldsymbol{\xi}_{\mathrm{via}}$ and the independent terminal joint $\vec{q}_{I,d}$. The dependent terminal joints are resolved analytically via $\vec{h}_D^{-1}$, and the full trajectory is represented as a cubic spline. The minimum-time parameterization accounts for the pump-flow rate constraint.}
\label{fig:AlgorithmOverview}
\end{figure*}

From a planning perspective, forestry cranes present two interacting challenges that distinguish them from standard robotic manipulators. First, their performance is fundamentally limited by the hydraulic pump flow rate (PFR), which is shared across all actuated joints simultaneously. Unlike systems with independent joint velocity limits, the PFR constraint introduces a nonlinear, globally coupled restriction: the admissible velocity of any joint depends on the instantaneous flow demand of all others. Time-optimal planning therefore requires coordinating joint motion collectively, rather than optimizing joints independently. Second, forestry cranes are kinematically redundant, multiple joint configurations correspond to the same end-effector pose in task space. This redundancy, if properly exploited, offers additional degrees of freedom to improve the global flow allocation across joints. However, if the terminal configuration is fixed prior to planning, this opportunity is lost entirely.

Existing approaches~\cite{kalmari:2014,kalmari:2017,dhakate:2022,morales:2014,hera:2021,ayoub:2024,spinelli:2025} address these challenges to varying degrees but do not jointly optimize trajectory duration under the hydraulic pump-flow constraint. The closest prior work, the family of VP-STO-based planners~\cite{ecker:2025,ecker:ifac:2025,ecker:2026}, explicitly integrates the PFR constraint and generates near-time-optimal trajectories, significantly outperforming RRT*-based approaches. 

Despite these advances, a fundamental limitation persists across all existing VP-STO-based planners: the terminal joint configuration must be fixed before trajectory optimization begins. This requires selecting a specific inverse kinematics solution a priori, before any knowledge of the required motion through the environment is available. For a redundant system under a globally coupled PFR constraint, this pre-commitment is particularly costly. The choice of terminal configuration strongly influences how hydraulic flow must be distributed among joints throughout the entire motion. Yet this distribution depends on the trajectory itself, which in turn depends on obstacle geometry, start configuration, and the joint coordination induced by the PFR constraint. Prematurely fixing the terminal configuration therefore eliminates the planner's ability to adapt joint coordination to the task, often forcing suboptimal flow distributions that increase total trajectory duration.

This paper addresses this limitation by introducing TSC-VP-STO, a task-space-constrained extension of VP-STO for forestry crane motion planning. Rather than prescribing a fixed terminal joint configuration, TSC-VP-STO enforces a terminal constraint directly in task space and treats the redundant degrees of freedom of the terminal configuration as additional free optimization variables. The trajectory and the terminal configuration are thus optimized jointly, allowing the planner to select a kinematically advantageous end configuration that improves hydraulic flow allocation across the full motion horizon. This yields more balanced pump utilization, faster joint coordination, and consistently shorter trajectory durations compared to the baseline.

The contributions of this paper are summarized in Fig.~\ref{fig:TSCVPSTOConcept} We present a general task-space-constrained formulation of VP-STO applicable to redundant manipulators under nonlinear actuation constraints. We introduce a structured configuration space decomposition into fixed, independent, and dependent joints, and derive an explicit reachability constraint for the forestry crane kinematics. We demonstrate through systematic numerical evaluation that TSC-VP-STO consistently reduces trajectory durations across all tested planning scenarios and via-point configurations, while maintaining comparable or lower total computation time. Finally, we validate the practical applicability of the proposed planner through deployment on a real forestry crane, including free navigation and a full log-loading cycle under near-time-optimal pump utilization.

\subsection{Related Work}

We structure the review around two topics directly relevant to the proposed approach: autonomous motion planning for large-scale hydraulic manipulators and redundancy resolution in kinematically redundant systems.

\paragraph{Autonomous motion planning for large-scale hydraulic manipulators}
Existing work on forestry crane automation spans model predictive path-tracking control with sway damping~\cite{kalmari:2014,kalmari:2017}, trajectory planning from pre-recorded human demonstrations~\cite{morales:2014}, learning-from-demonstration via dynamic motion primitives~\cite{hera:2021}
and reinforcement learning for end-effector tracking~\cite{dhakate:2022}.
At the system level, RRT*-based planners have been integrated into fully autonomous pipelines for log loading~\cite{ayoub:2024} and material handling~\cite{spinelli:2025}. However, none of these approaches jointly optimize trajectory duration under the hydraulic pump-flow constraint.

Jebellat and Sharf~\cite{jebellat:2023,jebellat:2024} address sway damping and waypoint tracking through dynamic programming, but do not explicitly optimize for execution time or account for the pump flow rate constraint.
Song and Sharf~\cite{song:2020} consider time-optimal motion planning with zero-moment point stability constraints for timber manipulation, but similarly do not incorporate the hydraulic pump-flow rate.
The most directly relevant prior work is the family of VP-STO-based planners introduced in~\cite{ecker:2025}, which explicitly integrates the pump-flow rate constraint into the time-optimal trajectory optimization and demonstrates near-time-optimal performance operating at the pump saturation limit and outperforming classical RRT* based approaches \cite{ayoub:2024,spinelli:2025}. This approach has subsequently been extended to account for pendulum sway dynamics~\cite{ecker:ifac:2025}, Euclidean Distance Field-based collision avoidance for slender robotic links~\cite{ecker:iros:2025}, and integrated with a collision-free sway-damping model predictive controller for closed-loop real-world deployment~\cite{ecker:2026}. The present work builds directly on this line of research and inherits its treatment of the pump-flow rate constraint.

\paragraph{Kinematic redundancy resolution}
Redundancy resolution, the problem of selecting among the infinite joint configurations that satisfy a given task-space constraint, is a classical topic in robot kinematics~\cite{siciliano:1990}. Common approaches include Jacobian null-space projection for local redundancy resolution, task augmentation, and optimization-based methods that exploit redundancy to satisfy secondary objectives such as joint limit avoidance, singularity avoidance, or manipulability maximization. In the context of motion planning, redundancy is typically resolved either at the inverse kinematics level prior to planning \cite{ecker:2025,ecker:ifac:2025,vu:2025}, or integrated into the planning problem itself \cite{ahuactzin:1999,keselman:2014,ferrentino:2021}. Pre-resolving redundancy decouples the IK and planning problems but eliminates the planner's ability to select terminal configurations that are advantageous for the planned motion. This limitation is well recognized in the manipulation literature~\cite{ahuactzin:1999,keselman:2014,ferrentino:2021}, and solutions for global planners such as RRT* and PRM have been proposed \cite{behnisch:2010,keselman:2014,ahuactzin:1999}. However, these approaches have not been extended to VP-STO-based planners for hydraulic crane planning under pump-flow constraints. This gap is particularly important because VP-STO has been shown to outperform classical sampling-based planners in terms of trajectory duration \cite{ecker:2025}, yet it does not explicitly account for the interaction between kinematic redundancy and the globally coupled actuation constraint, which is especially pronounced in pump-flow-limited hydraulic systems.

\section{Task-Space Constrained VP-STO}
We consider the mapping from joint space coordinates $\vec{q}\in\mathbb{R}^n$ to task-space coordinates
\begin{align}\label{eq:OutputMapping}
    \vec{y}=\vec{h}(\vec{q})\in\mathbb{R}^m\Comma
\end{align}
for a redundant manipulator, i.e. $m<n$. The planning task considered in this work can be formulated as an optimal control problem of the form
\begin{subequations}\label{eq:VPSTOOCP}
\begin{eqnarray}
    & \min\limits_{\vec{q}(\cdot),T} & J(\vec{q},\dot{\vec{q}},\ddot{\vec{q}},T)\\
    & \text{s.t.}             & \vec{q}(0)=\vec{q}_0\Comma\ \ \dot{\vec{q}}(0)=\vec{0}\\
    &                         & \vec{y}(T)=\vec{y}_d\Comma\ \ \dot{\vec{q}}(T)=\vec{0}\\
    &                         & \dot{\vec{q}}_{\mathrm{min}}\leq\dot{\vec{q}}(t)\leq\dot{\vec{q}}_{\mathrm{max}}, \ Q(t)\leq Q_{\mathrm{max}}\label{eq:VPSTOVeloConstr}\\
    &                         & \ddot{\vec{q}}_{\mathrm{min}}\leq\ddot{\vec{q}}(t)\leq\ddot{\vec{q}}_{\mathrm{max}},\ \vec{q}(t)\in\mathcal{C}_{\mathrm{free}}\label{eq:VPSTOAccConstr}\Comma
\end{eqnarray}
\end{subequations}
with the PFR constraint $Q(t)\leq Q_{\mathrm{max}}$ and the collision constraint $\mathbf{q}(t)\in\mathcal{C}_{\mathrm{free}}$.
In the context of this work, we set $J = T+J$, i.e., the objective is to minimize the total trajectory duration.
In contrast to the original formulation, which enforces a fixed terminal configuration $\vec{q}(T)=\vec{q}_d$, we only require the terminal task-space constraint $\vec{y}(T)=\vec{y}_d$, allowing the optimizer to select a kinematically advantageous final joint configuration. The main idea of the algorithm is summarized in Fig.~\ref{fig:AlgorithmOverview}.

\subsection{Original Baseline VP-STO}
This section briefly reviews the core concepts of the original Via-Point-based Stochastic Trajectory Optimization (VP-STO) algorithm \cite{janakowski:2023}. As previously noted, the algorithm solves the motion planning problem \eqref{eq:VPSTOOCP} given a fixed target configuration in joint space, $\vec{q}(T)=\vec{q}_d$. The trajectory is parameterized by $N$ via-points
\begin{align}\label{eq:VPSTO:ViaPoints}
    \boldsymbol{\xi}_{\mathrm{via}}=\begin{bmatrix}
        \vec{q}_1\\
        \vdots\\
        \vec{q}_N
    \end{bmatrix}
\end{align}
which are connected using cubic spline interpolations. This is formulated as the following minimum-acceleration optimization problem
\begin{subequations}\label{eq:SplineOptProblem}
\begin{eqnarray}
    &\min\limits_{\vec{q}(\cdot)} & \int_0^1\transpose{\vec{q}^{\prime\prime}(s)}\vec{q}^{\prime\prime}(s)\dd s\\
    &\text{s.t.} &\vec{q}(s_k)=\vec{q}_k\Comma\ \ k=1,\dots,N\\
    &            &\vec{q}(0) = \vec{q}_0\Comma\ \ \vec{q}^{\prime}(0)=\vec{0}\\
    &            &\vec{q}(1) = \vec{q}_d\Comma\ \ \vec{q}^{\prime}(1)=\vec{0}\FullStop
\end{eqnarray}
\end{subequations}

The solution of (\ref{eq:SplineOptProblem}) is given by cubic splines and can be written as $\vec{q}(s)=\vec{\Phi}(s)\vec{w}$ with the weight vector
\begin{align}
    \vec{w}&=\begin{bmatrix}
        \boldsymbol{\xi}_{\mathrm{via}}\\
        \boldsymbol{\xi}_{\mathrm{bc}}
    \end{bmatrix}\Comma
\end{align}
where $\transpose{\boldsymbol{\xi}_{\mathrm{bc}}}=[\transpose{\vec{q}_0},\transpose{\vec{q}_d},\transpose{\vec{0}},\transpose{\vec{0}}]$ is the vector containing the boundary conditions.
A linear time parameterization $s=\frac{t}{T}$ is used to yield the trajectory $\vec{q}(t)$.  The final time $T$ is chosen as the minimum time that fulfils the inequality constraints \eqref{eq:VPSTOVeloConstr} and \eqref{eq:VPSTOAccConstr}. Hence, (\ref{eq:VPSTOVeloConstr}) and (\ref{eq:VPSTOAccConstr}) yields a finite number of constraints of the form $T\geq T_j(\boldsymbol{\xi}_{\mathrm{via}})$. The minimum endtime that ensures (\ref{eq:VPSTOVeloConstr}) and (\ref{eq:VPSTOAccConstr}) can be computed by $T^*(\boldsymbol{\xi}_{\mathrm{via}})=\max_jT_j(\boldsymbol{\xi}_{\mathrm{via}})$. This allows to rewrite (\ref{eq:VPSTOOCP}) to
\begin{align}
    \min_{\boldsymbol{\xi}_{\mathrm{via}}} J\Big(\vec{q}(\boldsymbol{\xi}_{\mathrm{via}}),\dot{\vec{q}}(\boldsymbol{\xi}_{\mathrm{via}}),\ddot{\vec{q}}(\boldsymbol{\xi}_{\mathrm{via}}),T^*(\boldsymbol{\xi_{\mathrm{via}}})\Big)\FullStop
\end{align}

With $\vec{\Phi}(s)=[\vec{\Phi}_{\mathrm{via}}(s),\vec{\Phi}_{\mathrm{bc}}(s)]$, a smoothness prior
\begin{subequations}\label{eq:SmoothnessPrior}
\begin{align}
    \vec{\Sigma}_{\mathrm{via}}&=\bigg(\int_{0}^{1}\transpose{\vec{\Phi}^{\prime\prime}_{\mathrm{via}}(s)}\vec{\Phi}^{\prime\prime}_{\mathrm{via}}(s)\mathrm{d}s\bigg)^{-1}\\
    \vec{\mu}_{\mathrm{via}}&=\vec{\Sigma}_{\mathrm{via}}\int_{0}^{1}\transpose{\vec{\Phi}^{\prime\prime}_{\mathrm{via}}(s)}\vec{\Phi}^{\prime\prime}_{\mathrm{bc}}(s)\mathrm{d}s\boldsymbol{\xi}_{\mathrm{bc}}
\end{align}
\end{subequations}
is used to initialize the sampling distribution $\vec{\xi}_{\mathrm{via}}\sim\mathcal{N}(\vec{\mu}_{\mathrm{via}},\vec{\Sigma}_{\mathrm{via}})$.

\subsection{Configuration Space Decomposition}
We decompose the configuration space as
\begin{align}
    \vec{q}=\begin{bmatrix}
        \vec{q}_I\\
        \vec{q}_F\\
        \vec{q}_D
\end{bmatrix}\in\mathcal{C}_I\times\mathcal{C}_F\times\mathcal{C}_D=\mathcal{C}
\end{align}
where $\vec{q}_I\in\mathcal{C}_I\subseteq\mathbb{R}^{n_I}$ are the independent joint coordinates, $\vec{q}_F\in\mathcal{C}_F\subseteq\mathbb{R}^{n_F}$ are the fixed, and $\vec{q}_D\in\mathcal{C}_D\subseteq\mathbb{R}^{n_D}$ are the dependent joints coordinates, with $m=n_F + n_D$ and $n=n_F+n_I+n_D$.

For a given task-space coordinate $\vec{y}\in\mathbb{R}^m$, the fixed coordinates are uniquely defined by the mapping
\begin{align}
    \vec{q}_F=\vec{h}_F^{-1}(\vec{y})\Comma
\end{align}
If $\vec{q}_F\neq\vec{h}_F^{-1}(\vec{y})$, the task space target $\vec{y}=\vec{h}(\vec{q}_I,\vec{q}_F,\vec{q}_D)$ cannot be reached for any choice of $\vec{q}_I$ and $\vec{q}_D$.

Furthermore, for a given $\vec{y}$ and independent coordinates $\vec{q}_I$, we assume that the dependent joints are uniquely determined by
\begin{align}
    \vec{q}_D=\vec{h}_D^{-1}(\vec{y},\vec{q}_I)\Comma
\end{align}
such that \begin{align}
    \vec{y}=\vec{h}\big(\vec{q}_I,\vec{h}_F^{-1}(\vec{y}),\vec{h}_D^{-1}(\vec{y},\vec{q}_I)\big)\FullStop
\end{align}
This decomposition isolates the independent degrees of freedom that remain available for optimization at the terminal configuration.
We assume that the mappings $\vec{h}_F^{-1}$ and $\vec{h}_D^{-1}$ are unique and a derivation for the specific timber crane use case is proposed in Section~\ref{sec:Application}.


\subsection{Augmented Problem Formulation}
To fully exploit the kinematic redundancy of the forestry crane, we extend the standard VP-STO framework by treating the redundant terminal configuration as a free parameter within the optimization. We define an augmented parameter vector
\begin{align}\label{eq:TSCVPSTO:ParameterVector}
    \vec{\theta}=\begin{bmatrix}
        \vec{\xi}_{\mathrm{via}}\\
        \vec{q}_{I,d}
    \end{bmatrix}\in\mathbb{R}^{N\cdot n+n_I}\Comma
\end{align}
that includes the independent components $\vec{q}_{I,d}$ of the terminal configuration $\transpose{\vec{q}_d}=[\transpose{\vec{q}_{I,d}},\transpose{\vec{q}_{F,d}},\transpose{\vec{q}_{D,d}}]$.

Since the fixed coordinates $\vec{q}_{F,d}=\vec{h}_F^{-1}(\vec{y}_d)$ can be computed directly from the task-space target, they are incorporated into the boundary condition vector $\vec{w}_{\mathrm{bc}}$. This ensures that $\vec{q}_{F,d}$ are explicitly accounted for in the smoothness prior \eqref{eq:SmoothnessPrior}. Consequently, the new boundary condition vector is defined as
\begin{align}\label{eq:TSCVPSO:BoundaryConditionVector}
\boldsymbol{\xi}_{\mathrm{bc}}=[\transpose{\vec{q}}_0,\transpose{\vec{q}_{F,d}},\transpose{\vec{0}},\transpose{\vec{0}}]\FullStop
\end{align}

With the augmented weight vector
\begin{align}
    \vec{w}=\begin{bmatrix}
        \vec{\theta}\\
        \vec{h}_D^{-1}(\vec{y}_d,\vec{q}_{I,d})\\
        \boldsymbol{\xi}_{\mathrm{bc}}
    \end{bmatrix}
\end{align}
the path is represented as the cubic spline $\vec{q}(s)=\vec{\Phi}(s)\vec{w}$, $s\in[0,1]$. By construction, the boundary condition vector \eqref{eq:TSCVPSO:BoundaryConditionVector} ensures that the terminal constraint $\vec{h}\big(\vec{q}(1)\big)=\vec{y}_d$ is satisfied.
Time parameterization and computation of the final time $T$  are performed as in the original formulations \cite{janakowski:2023,ecker:2025}.

\subsection{Reachability Constraint}
For a given independent joint configuration $\vec{q}_I$, we define the set of reachable task-space coordinates as
\begin{align}
    \mathcal{R}(\vec{q}_I)=\bigg\{\vec{y}\in\mathbb{R}^m\Big|\mathbf{h}_D^{-1}(\vec{y},\vec{q}_I)\text{ exists}\bigg\}\FullStop
\end{align}
Feasibility of the terminal configuration thus requires
\begin{align}
    \vec{y}_d\in\mathcal{R}(\vec{q}_{I,d})\FullStop
\end{align}
This constraint is enforced via penalty terms, analogous to joint limit and collision constraints. Consequently, an explicit characterization of $\mathcal{R}(\vec{q}_I)$ is required for implementation. A concrete implementation for the timber crane can be found in Section~\ref{sec:ReachabilityConstraint}.

\subsection{Optimization}
The augmented parameter vector $\vec{\theta}$ is optimized using the covariance matrix adaptation (CMA) update rule of VP-STO~\cite{janakowski:2023}. In each iteration, $N_{\mathrm{sample}}$ candidate solutions are sampled from the current distribution, evaluated in parallel, and the distribution parameters are updated based on the top-ranked samples. The solver simultaneously refines path shape and the terminal configuration, ensuring that the selected terminal state is not merely a valid inverse kinematics solution, but the specific configuration that minimizes the total trajectory duration $T^*(\vec{\theta})$. In our experiments, we use $N_{\mathrm{sample}}=50$ trajectory samples and $100$ evaluation points along the spline.

\begin{figure}
\centering
\includegraphics[scale=0.28]{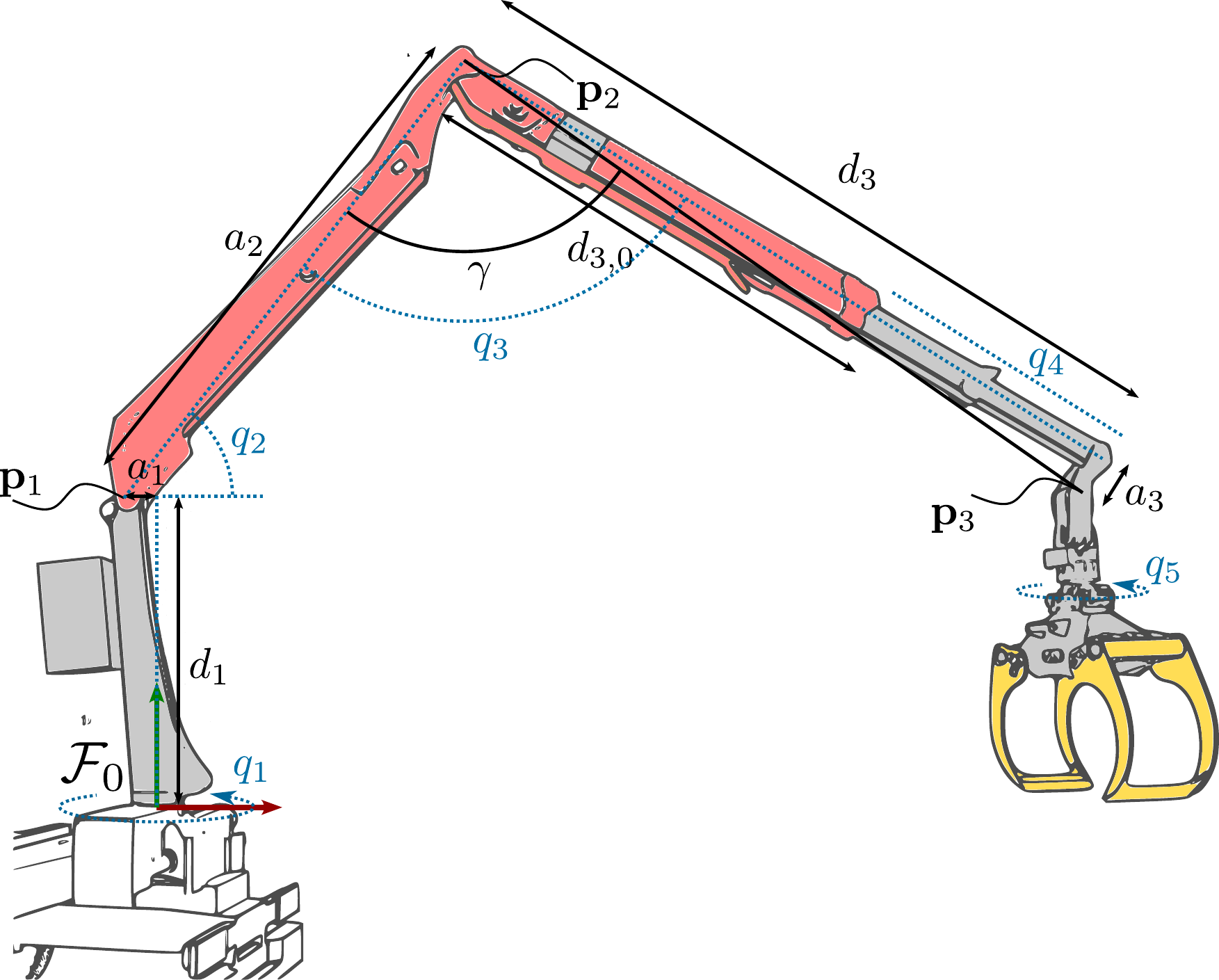}
\caption{Kinematic structure of the five-DOF forestry crane used for global motion planning. The joints $q_1$ (slewing), $q_2$ (main boom), $q_3$ (inner boom), $q_4$ (telescoping), and $q_5$ (gripper rotation) are shown. Points $\vec{p}_1$, $\vec{p}_2$, and $\vec{p}_3$ denote the relevant kinematic reference points used in the configuration space decomposition, and $\gamma$ is the angle used to compute the dependent joint $q_3$.}
\label{fig:KinematicChainDOF}
\end{figure}

\section{Forestry Crane Application}\label{sec:Application}
We apply TSC-VP-STO to the forestry crane depicted in Fig.~\ref{fig:TSCVPSTOConcept}, adopting the crane model from \cite{ecker:2025,ecker:ifac:2025}. The crane is modeled with five actuated DOFs, $\vec{q} = [q_1, \dots, q_5]^\mathrm{T}$. Passive pendulum dynamics at the tip are neglected at the global planning level and handled by the lower-level tracking controller~\cite{ecker:2025, ecker:2026}.

We define the task space coordinates
\begin{align}
    \vec{y}=\vec{h}(\vec{q})=\begin{bmatrix}
        \vec{p}_{\mathrm{tip}}\\
        \phi
    \end{bmatrix},
\end{align}
where $\vec{p}_{\mathrm{tip}}=[y_1,y_2,y_3]^{\mathrm{T}}\in\mathbb{R}^3$ denotes the Cartesian position of the tip, and $\phi\in[-\pi,\pi)$ is the yaw angle of the gripper expressed in frame~$\mathcal{F}_0$.  

\subsection{Pump flow rate}\label{sec:PumpFlowRate}
The displacements of the hydraulic cylinders denoted by $\vec{d}^{\mathrm{T}}=[{d}_1,\dots,{d}_{5}]\in\mathbb{R}^{5}$, are described as a function of $\vec{q}$ by $\vec{d}=\vec{h}_C(\vec{q})$. This directly results in the differential kinematics 
\begin{align}\label{eq:HydraulicVelocities}
    \dot{\vec{d}}&=\vec{J}_C(\vec{q})\dot{\vec{q}}=\frac{\partial\vec{h}_C(\vec{q})}{\partial\vec{q}}\dot{\vec{q}} \ ,
\end{align}
where $\dot{\vec{d}}\in\mathbb{R}^{5}$ are the velocities of the hydraulic actuators and $\dot{\vec{q}}$ are the joint velocities. The cylinders are supplied by a single pump, resulting in a pump flow rate (PFR) for the hydraulic system given by
\begin{align}\label{eq:TotalPumpFlow}
    Q(t) &= \sum_{l=1}^{5}A_l\big(\mathrm{sign}(\dot{d}_l)\big)|\dot{d}_l| \ ,
\end{align}
where $A_l(\cdot)$, $l=1,\dots,5$ are the direction-dependent effective areas of the cylinders. The PFR (\ref{eq:TotalPumpFlow}) that can be supplied by the hydraulic actuation system is limited by $Q_{\mathrm{max}}$. This yields the pump flow rate constraint (PFRC) $Q(t)\leq Q_{\mathrm{max}}$. For a detailed description on how to integrate this constraint into the endtime computation of VP-STO, the reader is referred to \cite{ecker:2025}.

\subsection{Configuration Space Decomposition}\label{sec:crane_decomp}
We choose the fixed joints as $\vec{q}_F=[q_1,q_5]^{\mathrm{T}}$, the independent joint as $q_I=q_2$, and the dependent joints as $\vec{q}_D=\transpose{[q_3,q_4]}$.

As illustrated in Fig.~\ref{fig:KinematicChainDOF}, given a desired tip position $\vec{p}_{\mathrm{tip},d}=[y_{1,d},y_{2,d},y_{3,d}]^{\mathrm{T}}$ and desired gripper orientation $\phi_d$, the fixed joints follow from
\begin{align}
    \vec{h}_F^{-1}(\vec{y})=\begin{bmatrix}
        \mathrm{atan2}(y_{2,d},\,y_{1,d})\\[4pt]
        \mathrm{atan2}(y_{2,d},\,y_{1,d})-\phi_d
    \end{bmatrix}.
\end{align}

To derive the mapping from the independent to the dependent joints, we express all relevant quantities in the frame $\mathcal{F}_0$, which is fixed to the slewing column and does not depend on~$q_1$.  
The points $\vec{p}_1$ and $\vec{p}_2$ in Fig.~\ref{fig:KinematicChainDOF} expressed in $\mathcal{F}_0$ read
\begin{align}
    \vec{p}_1&=\begin{bmatrix}
        -a_1\\
        d_1
    \end{bmatrix}, &
    \vec{p}_2&=\begin{bmatrix}
        a_2\cos q_2 - a_1\\[2pt]
        a_2\sin q_2 + d_1
    \end{bmatrix},
\end{align}
where $q_2=q_I$ is known.  
The tip position expressed in $\mathcal{F}_0$ is
\begin{align}
    \vec{p}_3=\begin{bmatrix}
        \sqrt{y_{1,d}^2+y_{2,d}^2}\\
        y_{3,d}
    \end{bmatrix}.
\end{align}
From Fig.~\ref{fig:KinematicChainDOF}, the quantity $d_3$ satisfies
\begin{align}\label{eq:Crane:d3}
    d_3 = \sqrt{\|\vec{p}_3-\vec{p}_2\|^2 - a_3^2}.
\end{align}
Because the telescope arm has a known fixed base length~$d_{3,0}$, the joint~$q_4$ follows as
\begin{align}
    q_4 = d_{3} - d_{3,0}.
\end{align}

The angle $\gamma$ shown in Fig.~\ref{fig:KinematicChainDOF} can be computed from $\vec{p}_1$, $\vec{p}_2$, and $\vec{p}_3$ via
\begin{align}
    \gamma = \arccos\bigg(
        \frac{(\vec{p}_3-\vec{p}_2)^\mathrm{T}(\vec{p}_1-\vec{p}_2)}
        {\|\vec{p}_3-\vec{p}_2\|\|\vec{p}_1-\vec{p}_2\|}
    \bigg),
\end{align}
which yields the dependent joint angle
\begin{align}
    q_3 = \gamma + \arctan\!\Big(\frac{a_3}{d_3}\Big).
\end{align}

Thus, the dependent joints can be obtained from the desired task-space configuration $\vec{y}_d$ and the independent joint $q_I=q_2$:
\begin{align}
    \vec{h}_D^{-1}(\vec{y},q_I)
    &= \begin{bmatrix}
        \gamma + \arctan\!\Big(\dfrac{a_3}{d_3}\Big)\\[6pt]
        d_3 - d_{3,0}
    \end{bmatrix}.
\end{align}

\begin{figure}
 \centering
 \adjustbox{trim=0cm 0cm 0cm 0cm,clip}{\includegraphics[scale=0.35]{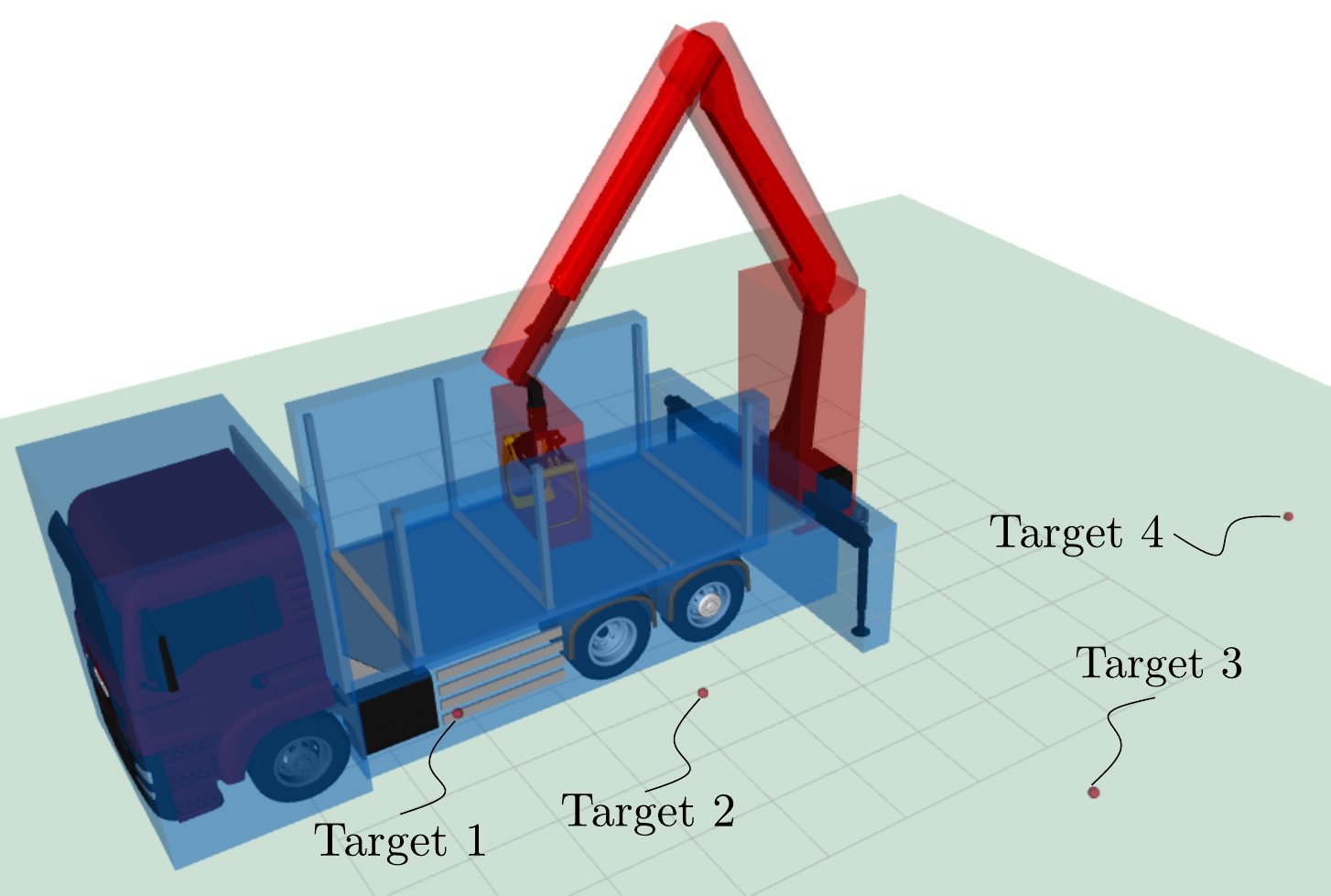}}
 \caption{Different task space targets used for numerical evaluation and comparison in Fig.~\ref{fig:DurationMeasurements}\label{fig:SimSetup}.}
 \label{fig:KinematicChain2D}
 \end{figure}
 
\begin{figure*}
    \centering
\adjustbox{trim=1.8cm 0.2cm 1.4cm 0.3cm,clip}{\includegraphics[scale=0.72]{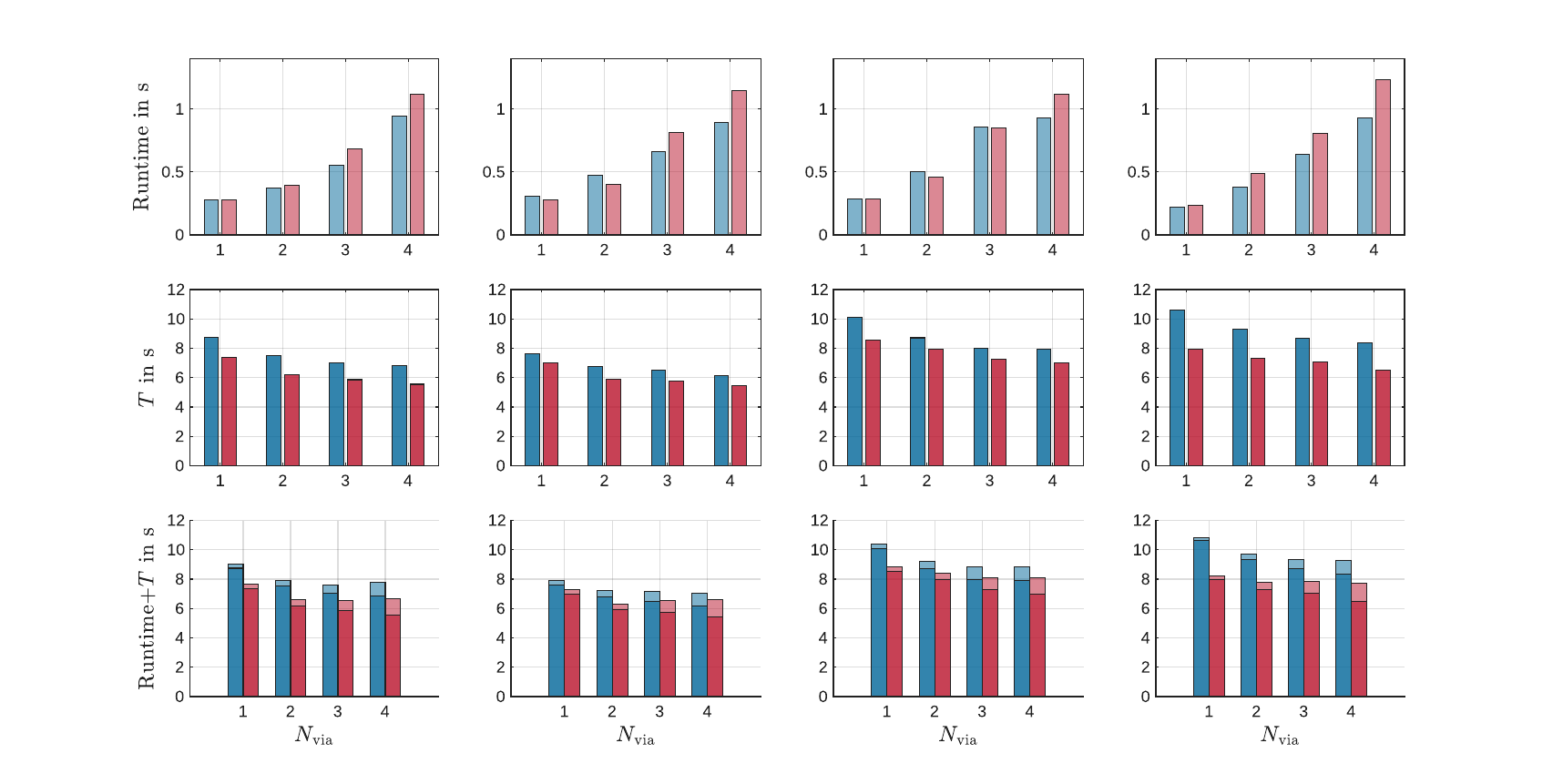}}
    \vspace{-6mm}
    \caption{Numerical comparison of TSC-VP-STO (red) and VP-STO (blue) across four planning targets and varying numbers of via-points $N_{\mathrm{via}}$. \textit{Top}: Mean planner runtime. \textit{Middle}: Mean trajectory duration $T$. \textit{Bottom}: Total execution time (runtime + trajectory duration). TSC-VP-STO consistently achieves shorter trajectory durations and lower total execution times across all scenarios.}
    \label{fig:DurationMeasurements}
\end{figure*}

 \begin{figure}
     \centering
     \includegraphics[scale=0.60]{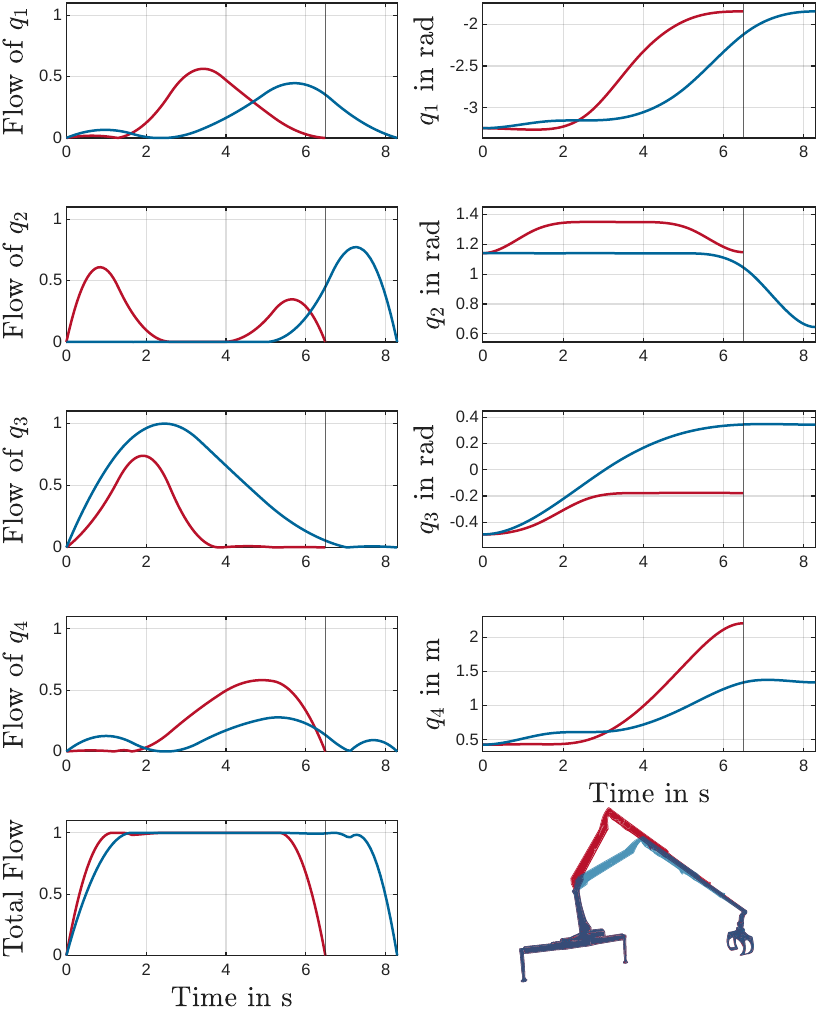}
     \vspace{-3mm}
     \caption{Representative sample trajectory for Target~4. \textit{Left}: Contribution of each joint to the total hydraulic pump flow rate, normalized by $Q_{\mathrm{max}}$. TSC-VP-STO (red) achieves a more balanced flow distribution than VP-STO (blue), with $q_3$ completing its motion early to free pump capacity for the remaining joints. \textit{Right}: Corresponding joint-space trajectories and target configurations.}
     \label{fig:SampleTrajectory}
 \end{figure}
\subsection{Reachability Constraint}\label{sec:ReachabilityConstraint}
From \eqref{eq:Crane:d3}, it follows that no solution exists whenever $\|\vec{p}_3-\vec{p}_2\|<a_3$.  
This has a clear physical interpretation: the tip position $\vec{p}_3$ must always remain at least the offset~$a_3$ away from $\vec{p}_2$ (ignoring joint limits, which are addressed separately).  
Accordingly, the reachable set for a given value of the independent joint $q_I=q_2$ is
\begin{align}
    \mathcal{R}(q_I)
    = \Big\{\vec{y}\in\mathbb{R}^4 : \|\vec{p}_3-\vec{p}_2\|\geq a_3\Big\}.
\end{align}
Joint limits are neglected in the definition of the reachable set, since they are addressed separately in the algorithm.

\section{Results}
We evaluate our algorithm on the forestry crane visualized in Fig.~\ref{fig:SimSetup}. We measure planning performance across four different task-space targets and varying numbers of via-points $N_{\mathrm{via}}$, comparing runtimes and overall trajectory duration against the classical VP-STO baseline~\cite{ecker:2025}. We then analyze the differences on a representative sample scenario to illustrate the coordination strategy employed by TSC-VP-STO. Finally, we deploy the planner on the real machine and demonstrate its practical applicability in a log-loading cycle.

\begin{figure*}
\centering
\begin{minipage}[t]{0.42\textwidth}
    \vspace{0pt}
    \includegraphics[trim=0cm 0.5cm 0cm 0cm, clip, width=\linewidth]{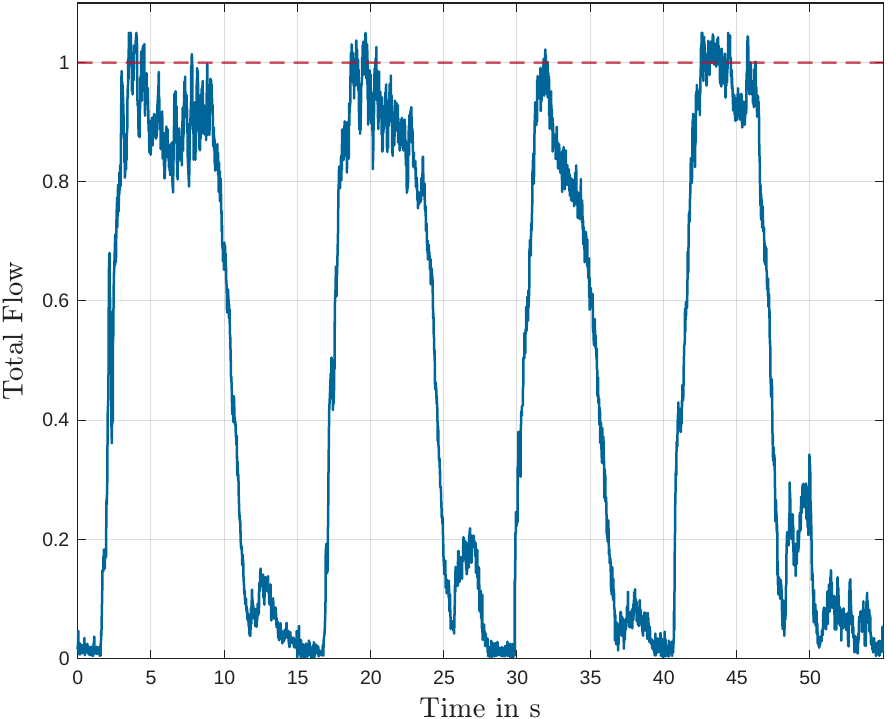}
\end{minipage}
\begin{minipage}[t]{0.505\textwidth}
    \vspace{0pt}
    \includegraphics[width=\linewidth]{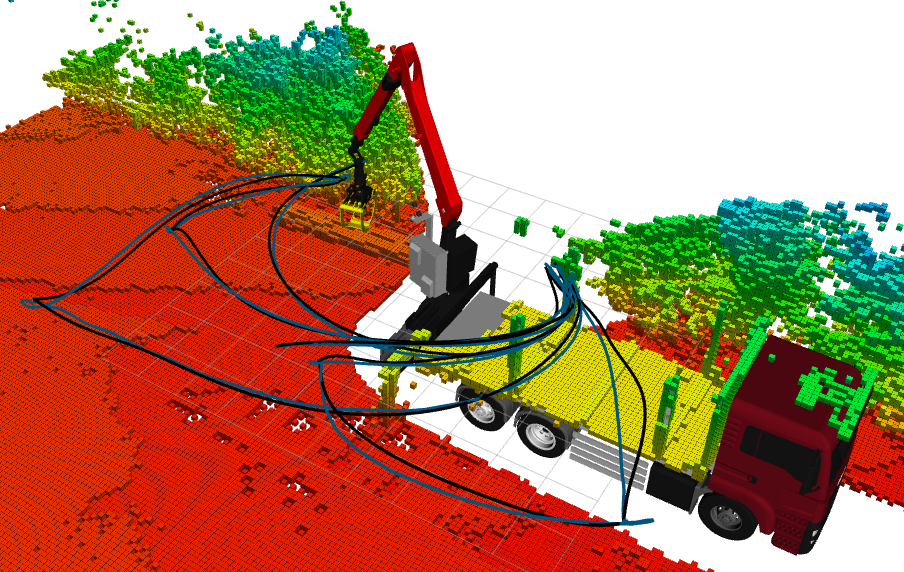}
\end{minipage}
    \centering
\begin{minipage}[t]{0.42\textwidth}
    \vspace{0pt}
    \includegraphics[width=\linewidth]{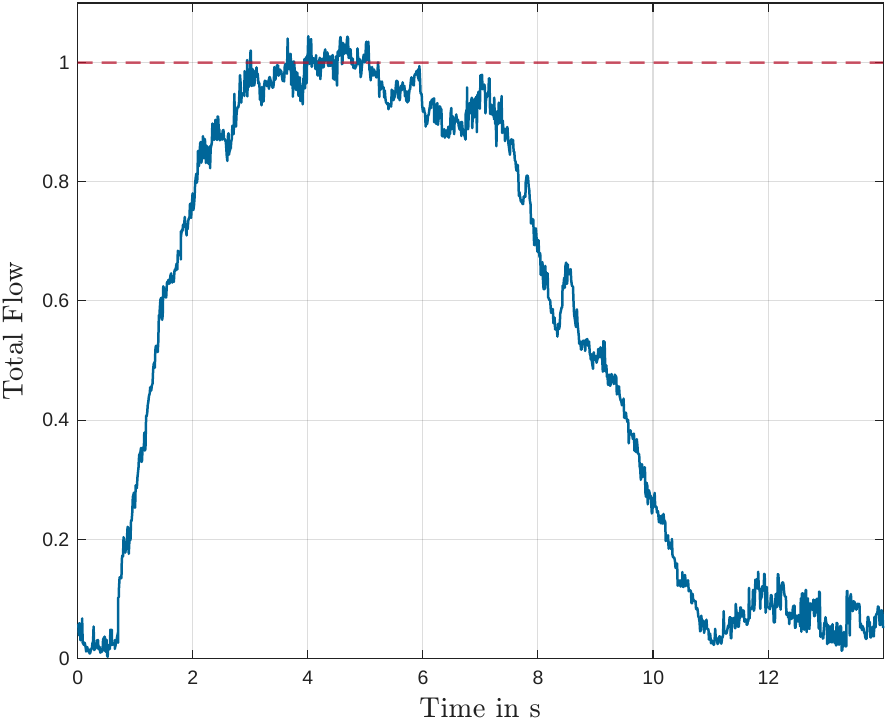}
\end{minipage}
\begin{minipage}[t]{0.505\textwidth}
    \vspace{0pt}
    \includegraphics[width=\linewidth]{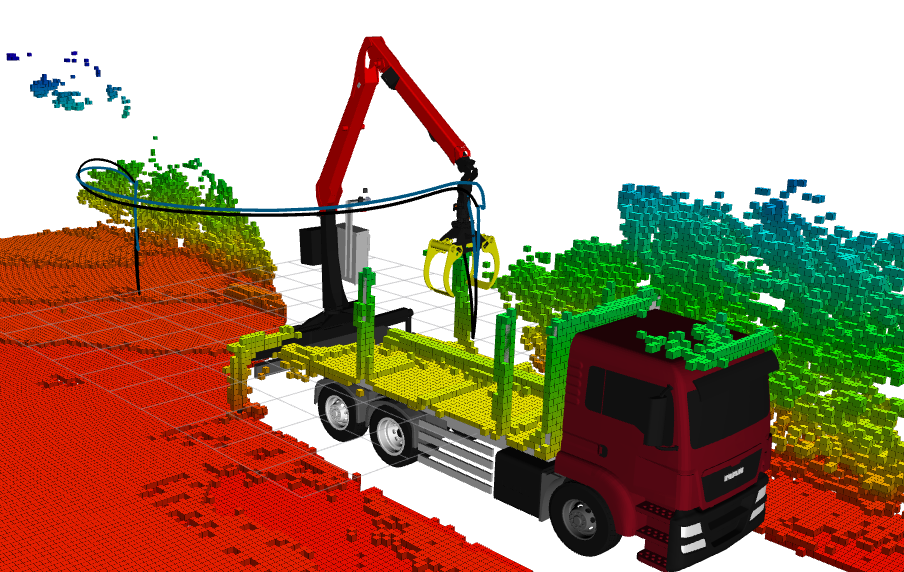}
\end{minipage}
\caption{Real-world crane experiments. \textit{Top}: Empty-gripper motions. \textit{Bottom}: Full log-loading cycle. \textit{Left}: Estimated hydraulic pump flow rate $Q_{\mathrm{pump}}\in[0,1]$, normalized by $Q_{\mathrm{max}}$; the dashed line indicates the flow rate limit. \textit{Right}: Planned tip trajectories (black) and executed tip trajectories (blue). The crane consistently operates near the pump flow rate limit, confirming near-time-optimal behavior.}
\label{fig:Experiment}
\end{figure*}
\subsection{Numerical Comparison}
\begin{table}
    \centering
    \caption{Average reduction of trajectory durations for 100 randomly selected task-space targets}\label{tab:Reductions}
    \vspace{-2mm}
    \begin{tabular}{||c||c||}
        \hline
         \multicolumn{1}{||c|}{$N_{\mathrm{via}}$} & \textbf{Avg. Reduction in \%}\\
         \hline\hline
         1 & 15.28\\
         2 & 13.11\\
         3 & 12.87\\
         4 & 12.37\\
        \hline
    \end{tabular}
\end{table}
We evaluate the proposed TSC-VP-STO against the baseline VP-STO introduced in \cite{ecker:2025} with the minimum norm IK strategy from \cite{vu:2025}. Both algorithms are tested with 100 evaluation points and 50 trajectory samples, a configuration that provides a good trade-off between runtime and trajectory quality. All trajectory evaluations are parallelized on eight CPU cores.

First, we evaluate the reduction of trajectory durations for 100 randomly selected task space targets, which are summarized in Table~\ref{tab:Reductions} . We found that on average our approach finds trajectories which are faster by 12-15\% than the baseline approach.

Fig.~\ref{fig:DurationMeasurements} compares both planners for different numbers of via-points $N_{\mathrm{via}}$ and the targets shown in Fig.~\ref{fig:SimSetup}. The figure shows (top) the mean planner runtime, (middle) the mean trajectory duration $T$, and (bottom) the total execution time, defined as the sum of runtime and trajectory duration.
This illustrates that our proposed TSC-VP-STO consistently reduces trajectory duration across all targets. By jointly optimizing the trajectory and the final joint configuration, the planner can exploit kinematic redundancy to minimize execution time in task space. Even with fewer via-points, TSC-VP-STO achieves faster trajectories than the baseline in most cases. Although its higher-dimensional search space and the lack of prior knowledge about a collision-free final configuration slightly increase optimization runtime, the overall total execution time remains lower than that of the classical VP-STO. Hence, TSC-VP-STO achieves better overall time-efficiency.

Since TSC-VP-STO reaches comparable or better trajectory durations with fewer via-points, it can reduce the number of decision variables while maintaining high-quality solutions, yielding shorter computation times overall.

For a more detailed analysis, Fig.~\ref{fig:SampleTrajectory} illustrates a representative example for Target 4. A consistent pattern is that TSC-VP-STO deliberately completes one joint ($q_3$) significantly earlier than the others, freeing pump capacity for the remaining joints. In contrast, VP-STO keeps $q_2$ nearly stationary initially and allocates most flow to $q_3$, leading to less balanced pump utilization. TSC-VP-STO distributes motion more evenly across joints over time, yielding smoother flow profiles and shorter total durations. The resulting terminal configurations for both planners are illustrated at the bottom right part of Fig.~\ref{fig:SampleTrajectory}.

\subsection{Real-World Implementation}
The proposed TSC-VP-STO planner was integrated into the autonomous control stack of a real-world forestry crane, following the framework in \cite{ecker:2026}. In this architecture, TSC-VP-STO serves as the global planning module, while a collision-free MPC acts as the underlying tracking controller to suppress pendulum oscillations. Environmental perception is handled via a LiDAR-based mapping system that generates a Euclidean Distance Field (EDF) for collision detection \cite{ecker:iros:2025}.

The performance of the planner is evaluated through a series of continuous trajectory executions. Fig.~\ref{fig:Experiment} (upper right) illustrates the end-effector motions, comparing the planned paths (black) with the actual crane movement (blue) using an empty gripper. The tracking performance remains high despite the high-speed nature of the motions. The larger deviation observed in the rightmost trajectory results from the MPC responding to collisions induced by the pendulum dynamics, which are not considered by the global planner.To verify the time-optimality of the generated plans, the estimated hydraulic pump flow rate is shown in the upper left of Fig.~\ref{fig:Experiment}. As direct measurement of the flow rate was unavailable, it was reconstructed from joint coordinate and velocity estimates.
The data reveals that the crane consistently operates at or near the defined flow rate limits during peak motion phases. This saturation confirms that the TSC-VP-STO planner effectively exploits the machine's actuation limits to produce near-time-optimal trajectories.

Finally, the system was validated in a full log-loading cycle, demonstrating that the task-space constrained approach is robust and applicable to real-world manipulation tasks. The results are illustrated at the lower part of Fig.~\ref{fig:Experiment}.

\section{Conclusion}
This paper presented TSC-VP-STO, a task-space-constrained extension of VP-STO for time-optimal motion planning on kinematically redundant forestry cranes under hydraulic pump-flow constraints. By replacing the fixed terminal joint-space constraint with a task-space constraint and jointly optimizing the trajectory and the redundant degrees of freedom of the terminal configuration, TSC-VP-STO enables the planner to adapt end configurations to the environment-dependent motion and hydraulic flow allocation.
We formalized the approach through a configuration space decomposition into fixed, independent, and dependent joints, and derived an explicit reachability constraint for the forestry crane kinematics. Experimental evaluations across multiple planning targets and via-point configurations consistently demonstrated reduced trajectory durations and improved pump-flow utilization compared to the VP-STO baseline. Real-world deployment on a forestry crane, including a full log-loading cycle, confirmed the practical applicability and near-time-optimal behavior of the proposed planner.

A current limitation is the assumption that the reachable set $\mathcal{R}(q_I)$ can be characterized in closed form. Extending to more complex kinematic chains may require numerical reachability estimation. Furthermore, the increased dimensionality of the search space due to the additional redundant terminal DOF may affect convergence for systems with higher degrees of redundancy.

\bibliographystyle{IEEEtran}
\bibliography{refs}

\end{document}